\documentclass[10pt,twocolumn,letterpaper]{article}

\usepackage[accsupp]{axessibility}

\usepackage[lined,ruled,linesnumbered]{algorithm2e}
\usepackage{animate} 

\usepackage{dsfont}
\usepackage[dvipsnames]{xcolor}
\usepackage[T1]{fontenc}

\usepackage{bmpsize}
\usepackage{epsfig}
\usepackage{float}
\usepackage{graphicx}
\usepackage[justification=raggedright]{caption}
\usepackage{lscape}
\usepackage{subcaption}
\usepackage{wrapfig}

\usepackage{array}
\usepackage{booktabs}                   
\usepackage{colortbl}
\usepackage{makecell}
\usepackage{multirow}
\usepackage{paralist}

\usepackage{amsmath}
\usepackage{pifont}
\usepackage{amssymb}
\usepackage{amsfonts}
\usepackage{bm}
\usepackage{mathtools}

\usepackage{comment}

\colorlet{dark-blue}{blue!50!black}
\colorlet{dark-cyan}{cyan!75!black}
\colorlet{dark-purple}{purple!50!black}
\colorlet{dark-red}{red!75!black}
\colorlet{dark-green}{green!80!black}
\colorlet{dark-orange}{orange!50!black}
\colorlet{dark-gray}{black!75}
\colorlet{light-gray}{black!30}
\definecolor{nice-red}{HTML}{E41A1C}
\definecolor{nice-orange}{HTML}{FF7F00}
\definecolor{nice-yellow}{HTML}{FFC020}
\definecolor{nice-green}{HTML}{39b54a}
\definecolor{nice-blue}{HTML}{0071bc}
\definecolor{nice-purple}{HTML}{984EA3}

\colorlet{verylight-gray}{black!10}
\definecolor{LightCyan}{rgb}{0.66,0.85,0.76}

\makeatletter
\DeclareRobustCommand\onedot{\futurelet\@let@token\@onedot}
\def\@onedot{\ifx\@let@token.\else.\null\fi\xspace}

\def\eg{\emph{e.g}\onedot} 
\def\ie{\emph{i.e}\onedot}

\def\etal{\emph{et al}\onedot}

\makeatother

\newcommand{\ignore}[1]{}   

\newcommand{\cmark}{\color{ForestGreen}{\ding{51}}}%
\newcommand{\xmark}{\color{BrickRed}{\ding{55}}}%

\newcommand{\mpage}[2]
{
\begin{minipage}{#1\linewidth}\centering
#2
\end{minipage}
}

\newcolumntype{L}[1]{>{\raggedright\let\newline\\\arraybackslash\hspace{0pt}}m{#1}}
\newcolumntype{C}[1]{>{\centering\let\newline\\\arraybackslash\hspace{0pt}}m{#1}}
\newcolumntype{R}[1]{>{\raggedleft\let\newline\\\arraybackslash\hspace{0pt}}m{#1}}

\newcommand{\modelname}{ShapeMove\xspace}
\newcommand{\betatoken}{[BETA]}

\definecolor{tabfirst}{rgb}{1, 0.7, 0.7} 
\definecolor{tabsecond}{rgb}{1, 0.85, 0.7} 
\definecolor{tabthird}{rgb}{1, 1, 0.7} 

\graphicspath{{figures}, {examples}}

\def\1n{\mathbf{1}_n}
\def\0{\mathbf{0}}
\def\1{\mathbf{1}}

\newcommand{\parens}[1]{\left(#1\right)}
\newcommand{\braces}[1]{\left\{#1\right\}}
\newcommand{\bracks}[1]{\left[#1\right]}
\newcommand{\modulus}[1]{\left\vert#1\right\vert}

\usepackage[pagenumbers]{cvpr} 

\definecolor{cvprblue}{rgb}{0.21,0.49,0.74}
\usepackage[pagebackref,breaklinks,colorlinks,allcolors=cvprblue]{hyperref}

\def\paperID{452} 
\def\confName{CVPR}
\def\confYear{2025}

\title{Shape My Moves: Text-Driven Shape-Aware Synthesis of Human Motions 
}

\author{%
    Ting-Hsuan Liao\textsuperscript{1, 2}\thanks{Work done while an intern at Adobe Research} \and
    Yi Zhou\textsuperscript{2} \and
    Yu Shen\textsuperscript{2} \and
    Chun-Hao Paul Huang\textsuperscript{2} \and
    Saayan Mitra\textsuperscript{2} \and
    Jia-Bin Huang\textsuperscript{1} \\
    \textsuperscript{1} University of Maryland, College Park, USA \and
    Uttaran Bhattacharya\textsuperscript{2} \\
    \textsuperscript{2} Adobe Research
}

\begin{document}

\twocolumn[{
\renewcommand\twocolumn[1][]{#1}
\maketitle
\begin{center}
    \centering
    \includegraphics[trim=0 0cm 0 0, clip,width=0.9\textwidth]{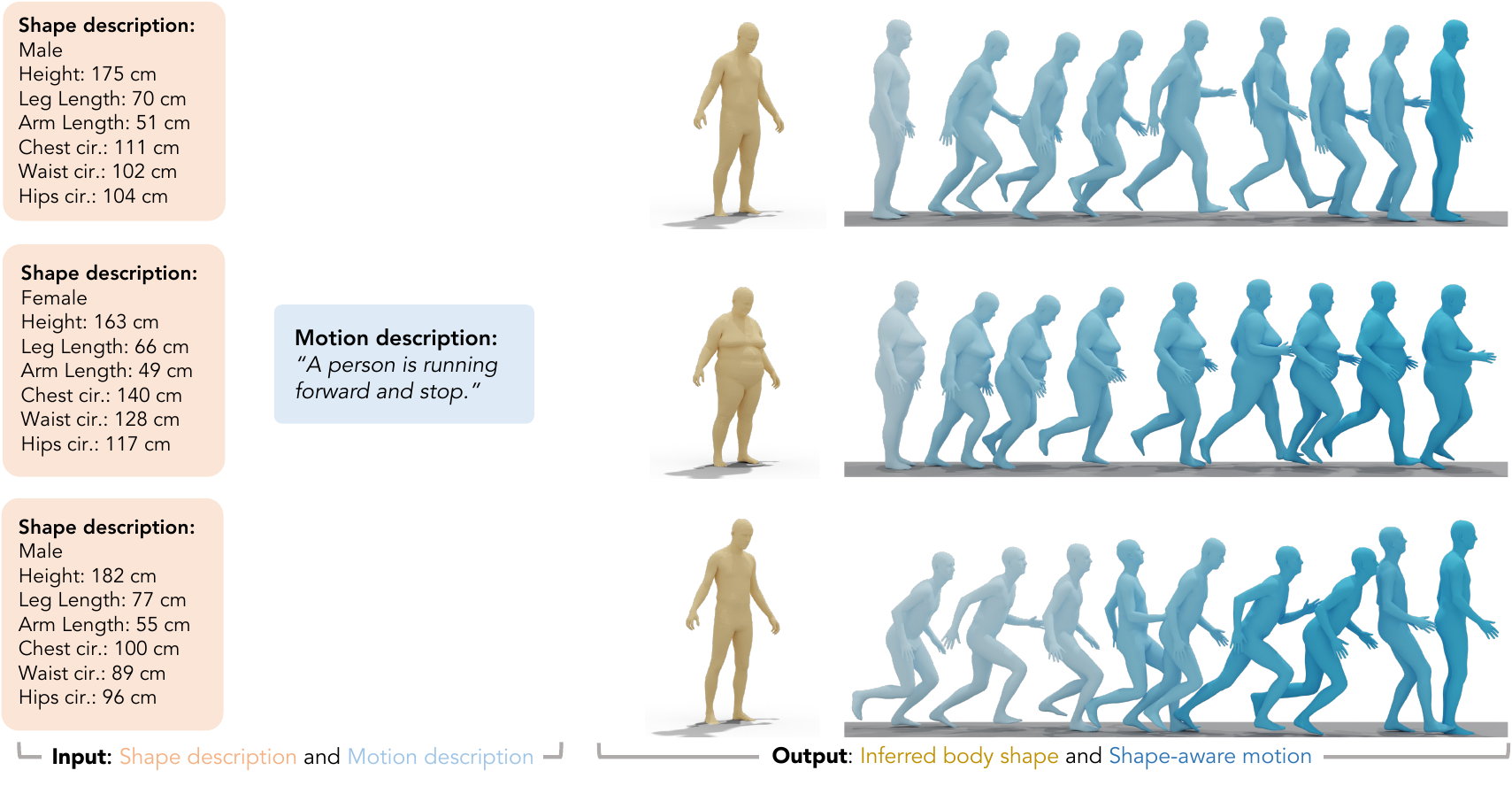}
    \captionof{figure}{\textbf{Text-Driven Shape-Aware Motion Synthesis.} The same motion performed by different body shapes can vary significantly, a realism aspect often overlooked in text-to-motion tasks. We propose a novel framework to integrate both shape and motion descriptions as input. Our framework synthesizes the shape parameters to reflect the described physical attributes, and injects them into motion synthesis to generate plausible shape-aware motions. The figure demonstrates the same \textit{running} motion synthesized across different body shapes.}
    \label{fig:teaser}
\end{center}
}]

\maketitle

\begin{abstract}
    We explore how body shapes influence human motion synthesis, an aspect often overlooked in existing text-to-motion generation methods due to the ease of learning a homogenized, canonical body shape. However, this homogenization can distort the natural correlations between different body shapes and their motion dynamics. Our method addresses this gap by generating body-shape-aware human motions from natural language prompts. We utilize a finite scalar quantization-based variational autoencoder (FSQ-VAE) to quantize motion into discrete tokens and then leverage continuous body shape information to de-quantize these tokens back into continuous, detailed motion. Additionally, we harness the capabilities of a pretrained language model to predict both continuous shape parameters and motion tokens, facilitating the synthesis of text-aligned motions and decoding them into shape-aware motions. We evaluate our method quantitatively and qualitatively, and also conduct a comprehensive perceptual study to demonstrate its efficacy in generating shape-aware motions. Project URL: {\normalsize \url{https://shape-move.github.io/}}.
\end{abstract}

\section{Introduction}
\label{sec:introduction}

Human motion synthesis is an expansive field with wide applications in avatar creation, robotics, and the gaming industry~\cite{starke2020local,unreal_engine,rokoko,motionbuilder}. Advances in both motion synthesis methods and language modeling techniques have, in recent times, led to the development of text-to-motion synthesis, with the aim of generating dynamic motions that are contextually aligned with textual descriptions.

Previous efforts in text-to-motion synthesis have explored techniques to integrate motion and language. Some approaches map motion and language into a shared latent space, and then sample motions from it based on text inputs~\cite{language2pose}. Others utilize diffusion models conditioned on learned features, such as CLIP, to enhance the quality and relevance of the generated motions~\cite{motionclip}. To overcome the difficulties of learning continuous motion features, quantizing motions into discrete tokens has emerged as an area of significant interest, which can then be predicted using transformers~\cite{t2mgpt} or fine-tuning large language models~\cite{motiongpt}. These advancements have led to improved realism and creative control over motion synthesis.

However, existing methods typically standardize motions by mapping to a canonical human body model. This results in homogenized motions across diverse body types and leaves out the unique attributes of individual body shapes. In reality, persons with different body shapes perform the same action with distinct physiological differences, as illustrated in Figure~\ref{fig:teaser}. Treating these distinct motions identically during motion synthesis leads to artifacts in subsequent motion transfer efforts~\cite{akber2023deeplearning}, often resulting in unrealistic motions and limitations on the body shapes on which the motions can be reliably retargeted.

Incorporating body shapes in motion synthesis is challenging due to the difficulties in both obtaining closed-form parameterization of motions by body shapes and learning the coarse-to-fine differences in motions due to individual body shapes in a data-driven manner. The challenges are exacerbated when attempting to merge continuous shape representations with quantized motion representations. To address these hurdles, we propose a quantization-based framework that seamlessly integrates continuous body shape data and thereby manages the information loss due to motion quantization. Consequently, our model not only predicts discrete motion tokens but also accurately captures continuous human body shape data, significantly enhancing the realism and individualism of synthesized motions.

Our model also leverages the capabilities of large language models to process complex language and predict sequential data, while handling continuous information through its latent space. This dual ability allows our model to predict continuous body shape parameters alongside discrete motion tokens. To maximize the benefits of this dual capability, our task-specific decoder utilizes discrete tokens and continuous shape parameter predictions as conditions to generate shape-aware human motions.

To summarize, our main contributions include:
\begin{itemize}
    \item End-to-end synthesis of both body shape parameters and shape-aware motions from input text descriptions.
    \item A novel framework to combine continuous shape and quantized motion data for efficient learning.
    \item Demonstrating efficacy through thorough quantitative and qualitative evaluations and human perceptual studies.
\end{itemize}

\section{Related Work}
\label{sec:rw}
We summarize existing research in the related areas of motion transfer, motion generation with and without body shape information, and motion generation from text inputs.

\begin{figure*}[t]
\centering
\includegraphics[trim=0.8cm 0cm 0.5cm 0, clip,width=\textwidth]
{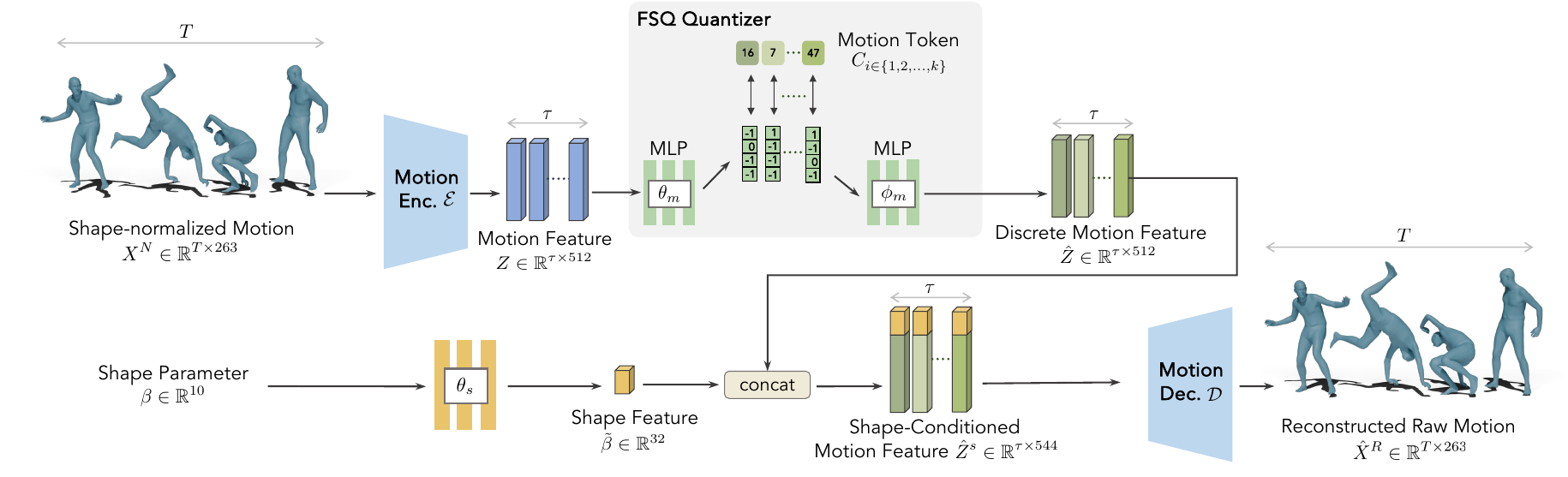}
\caption{
\textbf{Shape-Aware FSQ-VAE (SA-VAE) Overview.} SA-VAE is our quantization network learning to generate discrete motion tokens. Given a shape-normalized motion $X^N \in \mathbb{R}^{T \times D}$ of length $T$ and dimensionality $D$ ($= 263$ in our setup), we first encode the motion with the Motion Encoder $\mathcal{E}$ into a motion feature $Z \in \mathbb{R}^{\tau \times D}$, where $\tau$ represents a downsampling of $T$. We leverage the FSQ~\cite{fsq} quantizer to quantize $Z$, which gives a discrete feature $\hat{Z}$.
The $\text{MLP}_{\theta_{m}}$ and $\text{MLP}_{\phi_{m}}$ transform the features into the required code dimensions. To condition on the shape, we project the shape parameter $\beta$ with the Projector $P_{\theta_{s}}$ to align with $\hat{Z}$. We concatenate the shape feature 
 $\tilde{\beta}$ with $\hat{Z}$, then feed it into the Motion Decoder $\mathcal{D}$ to predict the reconstructed motion ${\hat{X}^R}$.
}
\label{fig:vqvae}
\end{figure*}

\paragraph{Motion Transfer.}
Motion transfer is a common strategy for applying motions from a source skeletal topology to a target one. While heuristic approaches based on inverse kinematics can resolve minor motion artifacts in the target skeleton, they cannot reliably reconcile motion artifacts due to topological mismatches between the source and the target~\cite{choi2000onlinemotion,unreal_engine,rokoko,motionbuilder}.
More direct approaches that require paired source and target motion data are limited by the practical challenges of obtaining such paired data, particularly as the topological differences between the source and the target skeletons increase~\cite{katsu2010animatingnonhumanoid,abdul2017motionstyleretargeting,zhou2020unsupervisedshape}.
Recent data-driven approaches relying on deep learning techniques~\cite{akber2023deeplearning} can alleviate the strict requirements of paired motion data~\cite{aberman2020skeletonaware,lee2023same}, and can even transfer motions between different modalities, such as from videos to 3D meshes~\cite{celikcan2015examplebased,aberman2019learningcharacter}, while plausibly adhering to physical constraints, such as foot contacts~\cite{basset2019contactpreserving,villegas2021contactaware}.
However, such approaches predominantly rely on skeletal topologies to transfer motions. They either cannot adapt to motion transfer between physiologically different body shapes or can only do so for a limited variety of body shapes and motions~\cite{wang2020neuralposetransfer,gomes2021ashapeawareretargeting,regateiro2022temporalshape,zhang2023skinnedmotion}.

\paragraph{Human Motion Generation with Body Shapes.}
We broadly categorize computational methods for human motion generation into \textit{shape-aware} --- those that consider body shapes in the generation process, and \textit{shape-free} --- those that generate motions for a normalized representation of the human body and use motion transfer techniques to apply the generated motions to different body shapes.
Owing to the simpler nature of its formulation, shape-free motion generation has been richly explored in computer vision and graphics. The underlying motion models are often developed with physically-based constraints~\cite{zell2017joint3dhuman,bergamin2019drecon,yuan2020dlow,yuan2020residualforce,shimada2020physcap,yuan2021simpoe,fussell2021supertrack,shimada2021neuralmonocular,yi2022physicalinertialposer,tripathi20233dhumanpose} and employ reinforcement learning frameworks for use within physics engines~\cite{schulman2015highdimensionalcc,schulman2017proximal,peng2017learninglocomotionskills,peng2018deepmimic,peng2018sfv,won2020ascalableapproach,makoviychuk2021isaac}. In addition to physical constraints, generative approaches are typically conditioned on guidance signals, such as past motions of the same person~\cite{komura2017arecurrentvariational} interactive motions of another person~\cite{ghosh2024remos,sui2025survey}, audio, including music~\cite{li2021aichoreographer,danceanyway} and speech~\cite{ginosar2019learningindividual,bhattacharya2021text2gestures}, text descriptions~\cite{language2pose,ghosh2021text_based,ghosh2021synthesisof}, and their various combinations~\cite{s2ag}. The scope of these approaches has expanded from deterministic, producing a fixed output for a given set of guidance signals~\cite{grenander1994representations,fragkiadaki2015recurrent,holden2016adeeplearning,ghosh2017learning,gopalakrishnan2019aneuraltemporal,aksan2019structuredprediction,guo2020action2motion,bao2022analyticdpm}, to stochastic, enabling more creative control over the generated motions~\cite{aliakbarian2020astochasticconditioning,he2022nemf,t2m,ren2023diffusionmotion}.
Generative frameworks such as GANs and VAEs, paired with the creation of standardized human representations~\cite{smpl} and large-scale benchmark datasets~\cite{AMASS}, have heralded the modern wave of motion generation methods. They have significantly improved the plausibility of generated motions on canonical skeletons~\cite{ahn2018text2action,barsoum2018hpgan,rempe2021humor}, and have been expanded to object~\cite{ghosh2023imos} and scene interactions~\cite{jiang2024scalingup}. More recent methods have leveraged versatile sequence modeling architectures such as transformers~\cite{lou2024multimodal} and robust generative techniques such as denoising diffusion models~\cite{dabral2023mofusion}.
Despite the remarkable progress in shape-free motion generation, their effective realism, when applied to diverse body shapes, remains susceptible to common motion artifacts, including self-intersections, foot contact errors, and unnatural pose articulations. Shape-aware pose models~\cite{shapy,semantify,bodyshapegpt} and motion generation methods~\cite{humos} have made progress in reducing this quality gap by explicitly incorporating body shape information in the generative process and employing physically-based constraints that support the shape information. Our work takes the next step in shape-aware motion generation by learning to generate motions from natural language descriptions of both shapes and motions in an end-to-end fashion.

\paragraph{Motion from Language Models.}
Motion generation from text descriptions has become especially active in recent human motion generation research, aided largely by the user-friendliness and creative utility of describing motions in free-form text via language models~\cite{language2pose,motionclip,actor,temos,tm2t,teach,athanasiou2023sinc,motionmamba}. Current state-of-the-art approaches typically employ diffusion-based~\cite{motiondiffuse,mld,mdm,priormdm,length-aware} networks or transformer-based~\cite{t2mgpt,motiongpt,momask} networks that map tokenized language features to motion features. In the quantization step, motion features are also often quantized and tokenized, following quantization paradigms such as the VQ-VAE~\cite{van2017neural}, to expedite the generative learning process while maintaining motion quality. We inject shape-awareness into the motion tokenization process through data-driven shape features and finite scalar quantization (FSQ)~\cite{fsq}, enabling us to map language tokens to combine shape and motion features for shape-aware motion generation from text.

\begin{figure*}[t]
\centering
\includegraphics[trim=0.3cm 0cm 0.1cm 0, clip,width=0.9\textwidth]
{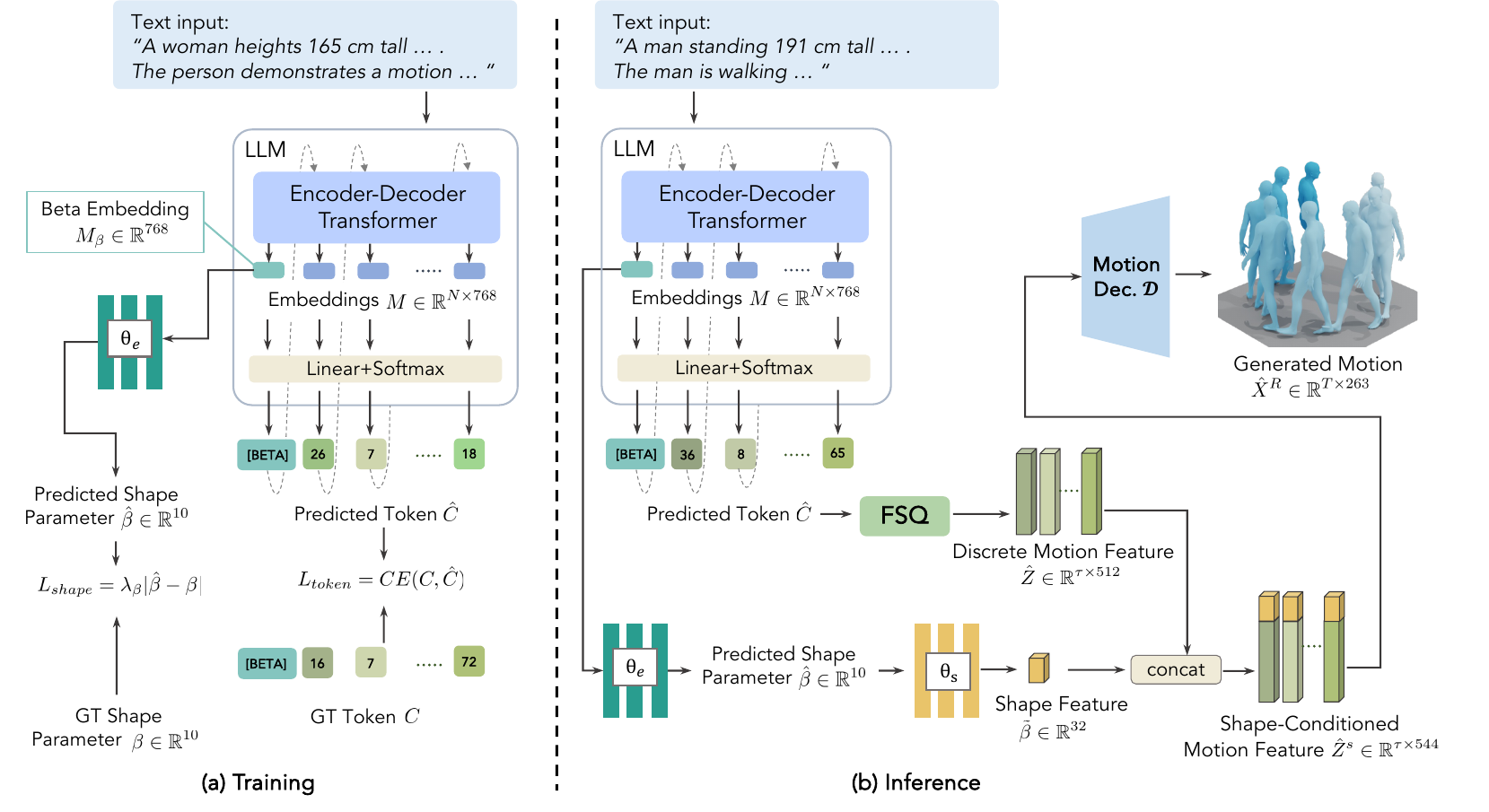}
\caption{
\textbf{\modelname~Overview.} In the training phase (a), the transformer network takes in the text inputs describing human motions and body shapes and predicts quantized motion tokens and the shape token \betatoken. The embedding for \betatoken~passes through the Projector \(P_{\theta_{e}}\) to predict the shape parameter \(\hat{\beta}\). We use cross-entropy loss for comparing ground truth tokens \(C\) with predicted tokens \(\hat{C}\), and \(L1\) loss for shape parameter to optimize the model. In the inference phase (b), our model predicts motion tokens \(\hat{C}\) and the shape parameter \(\hat{\beta}\) from text inputs. We de-quantize these tokens using FSQ, and project into shape parameters with Projector \(P_{\theta_{s}}\). We concatenate \(\hat{\beta}\) and \(\hat{C}\), and decode into the generated motion sequence with the Motion Decoder \(\mathcal{D}\). Our model effectively synthesizes shape parameters and shape-aware motions reflecting the physical form and actions described in the input text.
}
\label{fig:token}
\end{figure*}

\section{Method}
\label{sec:method}

We aim to generate both human body shapes and motions from natural language text descriptions. To this end, we first preprocess text-to-motion data to obtain shape descriptions and shape-aware motion data (Section~\ref{sec:data}). We propose a two-stage framework for shape-aware motion synthesis from text, consisting of a shape-aware FSQ-VAE (SA-VAE) and a shape-motion token predictor (\modelname). The SA-VAE efficiently quantizes shape-normalized motion into discrete tokens and reconstructs motions from shape parameters (Section~\ref{sec:vae}). \modelname generates both shape tokens and a sequence of motion tokens that accurately correspond to the text description (Section~\ref{sec:token_predict}).

\subsection{Data Preprocessing}
\label{sec:data}

Existing text-motion datasets only provide natural language labels to describe motions~\cite{t2m,kit}. To obtain specific text labels for body shape, we utilize the SMPL~\cite{smpl} model's shape parameters $\beta$, which control the body shape, and extract the measurements of six attributes to get a holistic view of the body: height, arm and leg lengths, and chest, waist, and hip circumferences. These attributes form our shape description, which we use as input for the model.

Since available text-motion datasets contain a limited diversity of characters, we propose generating additional body shapes for training. We utilize the Attributes to Shape (A2S) model of Shapy~\cite{shapy}, which can generate SMPL beta values from linguistic shape attributes to generate more body shapes in a controlled and plausible manner. We detail the pipeline for using additional shapes in Section~\ref{sec:vae}.

For the motion representation, we follow the preprocessing procedure of Guo \etal~\cite{t2m}, extracting joint positions from the SMPL layer and computing motion representations with shape $T\times D$, where $D$=263 is the data dimension and $T$ is the number of motion frames. Each motion representation contains root relative rotation and velocity, root height, joint locations, velocities, rotations, and foot contact labels. Notably, we skip the normalizing step of canonicalizing the skeletons to a homogeneous body shape. We denote this as the shape-aware motion $X^R$, and the corresponding shape-normalized version of the motion as $X^N$.

\subsection{Shape-Aware FSQ-VAE (SA-VAE)}
\label{sec:vae}

Quantizing motion into discrete tokens has led to significant progress in the text-to-motion synthesis~\cite{tm2t, t2mgpt, motiongpt}. Existing approaches quantize motion into discrete tokens and then predict sequences of motion tokens that align with the text input. However, the codebook capacities of these approaches are insufficient for reconstructing shape-variant motions ($X^R$). To address this limitation, we propose a new framework, Shape-Aware FSQ-VAE (SA-VAE), to effectively quantize the shape-normalized motions $X^N$ into discrete tokens and reconstruct shape-aware motions $\hat{X}^R$ by conditioning on shape information. We use $X^N$ instead of $X^R$ as the SA-VAE input as it enables the SA-VAE to disentangle the \textit{content} (given by the pose sequences) and the \textit{style} (given by the body shapes) components of the motions, focusing on learning tokens for the motion content and shape features for the motion style separately, and then combining them for efficient decoding into $X^R$.

As illustrated in Figure~\ref{fig:vqvae}, our SA-VAE comprises a motion encoder $\mathcal{E}$ and a motion decoder $\mathcal{D}$. To learn a more generalized motion codebook, the encoder $\mathcal{E}$ takes in a sequence of shape-normalized motion frames \(X^N = \bracks{x^N_1, x^N_2, \ldots, x^N_T}\), \(x^N_i \in \mathbb{R}^D\) and encodes them into motion features \(Z = \bracks{z_1, z_2, \ldots, z_\tau}\),  \(z_i \in \mathbb{R}^d\) and \(\tau\) representing a downsampling of \(T\). 

We utilize Finite Scalar Quantization (FSQ)~\cite{fsq} as our quantizer, given its superior codebook utilization without needing the additional regularization typically used in previous quantizers~\cite{t2mgpt}. We denote the quantized features as
\(
\hat{Z} = \mathrm{MLP}_{\phi_{m}}\parens{
\nabla\lfloor f\parens{
\mathrm{MLP}_{\theta_{m}}\parens{Z}}\rceil_{ste}}\), where \(f\) is a bounding function that confines the latent within a specific range, defined as \(f \lfloor L/2 \rfloor \odot \tanh(Z)\), and \(L \in \mathbb{Z}^\ell\) determines the width of the codebook for each dimension. We round the latents to integers with straight-through gradients \(\parens{\nabla\lfloor\cdot\rceil_{ste}}\). The MLPs, \(\mathrm{MLP}_{\theta_{m}}\) and \(\mathrm{MLP}_{\phi_{m}}\), transform the features into required code dimensions. 

Our motion decoder $\mathcal{D}$ decodes the discrete motion features 
\(\hat{Z} = \bracks{\hat{z}_1, \hat{z}_2, \ldots, \hat{z}_\tau}\), \(\hat{z}_i \in \mathbb{R}^d\) into shape-aware motions. To incorporate shape information, we extract the shape feature \(\tilde{\beta} = \mathrm{P}_{\theta_{s}}(\beta)\), where \(\mathrm{P}_{\theta_{s}}\) is an MLP-based projector. We concatenate the shape feature along the time dimension with the discrete motion feature to feed into $\mathcal{D}$. We denote the final reconstructed shape-aware motion as \(\hat{X} = \mathcal{D}\parens{\hat{Z} \bigoplus \tilde{\beta}}\), \(\bigoplus\) denoting concatenation.

\begin{table*}[t]
\centering
\caption{\textbf{Comparison with Baselines.} 
We evaluate the Penetrate, Float, Skate Ratio, and Bone Length Variances for our method and available baselines. For fair comparison, retrain baselines with shape-aware motions. The \textit{Shape Input Capability} column indicates methods that can incorporate both shape and motion descriptions --- a {\xmark} here suggests the corresponding text encoder cannot parse shape descriptions. \textit{Arbitrary Length} denotes results obtained without using ground-truth motion lengths. We compare with methods that share the same constraints as ours, capable of generating arbitrary motion lengths and accepting shape descriptions as input. Our method achieves the best or comparable results across the board. It particularly excels in the Penetrate metric, showcasing its adeptness at handling shape-aware motions.
$\colorbox[RGB]{255,179,179}{\phantom{a}}$ and $\colorbox[RGB]{255,217,179}{\phantom{a}}$ highlight the \colorbox[RGB]{255,179,179}{best} and \colorbox[RGB]{255,217,179}{second-best} results.
}
\resizebox{\textwidth}{!}{
\begin{tabular}{lcccccccccccc}
\toprule
\multirow{2.2}{*}{Methods} & \multirow{1.8}{*}{Shape Input} & \multirow{1.8}{*}{Arbitrary} & 
\multirow{1.8}{*}{Penetrate} & \multirow{1.8}{*}{Float} & \multirow{1.8}{*}{Skate } & \multirow{1.8}{*}{Bone Length} & 
\multicolumn{3}{c}{RPrecision$~\uparrow$} & \multirow{2.2}{*}{FID$~\downarrow$} & \multirow{2.2}{*}{MMDist$~\downarrow$} & \multirow{2.2}{*}{Diversity$~\downarrow$} \\
\cmidrule(lr){8-10}
& \multirow{1.0}{*}{Capability} & \multirow{1.0}{*}{Length} & {($cm$)$~\downarrow$} & {($cm$)$~\downarrow$} & {($\%$) $~\downarrow$} & {Variances$~\downarrow$}
& {Top1} & {Top2} & {Top3}  && \\
\midrule
Real         & $-$ & $-$ & $0.0$ & $0.0362 $ & $ 7.468$ & $ 0.0$ & $0.469 $ & $0.665 $ & $0.769 $ & $0.001$ & $3.217 $ & $0.000$ \\
SA-VAE (Recon.)          & $-$ & $-$ & $0.0289$ & $0.2090$ & $6.443$ & $0.623$  & $ 0.454$ & $0.645 $ & $0.749 $ & $0.125$ & $3.308 $ & $0.101$ \\
\midrule
TM2T   ~\cite{tm2t} & \xmark   &        \cmark & $0.1485 $ & $\textcolor{black}{0.2456} $ & $ 8.554 $ & $ 5.339$ & $0.374 $ & $0.559  $ & $0.673 $ & $1.671 $ & $3.843 $ & $0.937 $ \\ 
T2M ~\cite{t2m}  & \xmark   &        \cmark  & $\textcolor{black}{0.0939} $ & $0.6805 $ & $ \textcolor{black}{4.250} $ & $ 1.352$  & $0.408 $ & $0.592  $ & $0.697 $ & $1.230 $ & $3.597 $ & $0.430 $ \\ 

MLD ~\cite{mld} & \cmark   &        \xmark  & $0.3091 $ & $0.6558 $ & $ 9.313 $ & $ 2.695$ & $0.383 $ & $0.571  $ & $0.680 $ & $0.882 $ & $3.736 $ & $\textcolor{black}{0.020} $ \\ 

MotionDiffuse  ~\cite{motiondiffuse}& \cmark   &        \xmark & $0.2401 $ & $0.2703 $ & $ 7.710 $ & $ \textcolor{black}{0.138}$ & $\textcolor{black}{0.426} $ & $ \textcolor{black}{0.616}  $ & $ \textcolor{black}{0.723} $ & $0.563 $ & $\textcolor{black}{3.392} $ & $ 0.320 $ \\ 

MDM   ~\cite{mdm}  & \cmark   &        \xmark  & $ 0.1011$ & $ 1.7101$ & $ 8.523$ & $ 0.666$ & $0.317 $ & $0.490  $ & $0.599 $ & $0.461 $ & $4.180 $ & $0.320 $ \\
\midrule
T2M-GPT ~\cite{t2mgpt}& \cmark   &        \cmark  & $\cellcolor{tabsecond}0.1789 $ & $0.5241 $ & $ \cellcolor{tabsecond}6.162 $ & $ \cellcolor{tabsecond}1.176 $ & $\cellcolor{tabsecond}{0.394} $ & $\cellcolor{tabsecond}{0.576}  $ & $\cellcolor{tabsecond}{0.683} $ & $\cellcolor{tabsecond}{0.269} $ & $\cellcolor{tabsecond}{3.710} $ & $\cellcolor{tabsecond}{0.190} $ \\
MotionGPT ~\cite{motiongpt} & \cmark   &        \cmark & $ 0.6986$ & $ \cellcolor{tabfirst}{0.2245}$ & $ 7.889$  & $ 2.271 $ & $0.128 $ & $0.208  $ & $0.271$ & $1.020 $ & $7.055 $ & $ 0.389$ \\
\modelname ~(Ours) & \cmark   &        \cmark &       $ \cellcolor{tabfirst} 0.0268$       &    \cellcolor{tabsecond}$0.2658 $      &     $\cellcolor{tabfirst}{6.143} $ & $\cellcolor{tabfirst}{0.625} $ & $\cellcolor{tabfirst}{0.413}$ & $\cellcolor{tabfirst}{0.601}  $ & $\cellcolor{tabfirst}{0.705} $ & $\cellcolor{tabfirst}{0.198} $ & $\cellcolor{tabfirst}{3.533} $ & $\cellcolor{tabfirst}{0.117} $ \\
\bottomrule
\end{tabular}
}

\label{tab:physic_text2motion}
\end{table*}

\paragraph{Objective.} Following the approach of T2M-GPT~\cite{t2mgpt}, we use \(L^{\mathrm{smooth}}_1\) to compute the reconstruction loss \(L_r\) between the ground truth and the reconstructed motions to train the SA-VAE. Denoting \(X_{\mathrm{rot}}\) as the invariant rotation information of \(X\), we write the reconstruction loss as
\begin{equation}
L_r = L^{\mathrm{smooth}}_1\parens{X, \hat{X}} + \lambda_{\mathrm{rot}} L^{\mathrm{smooth}}_1\parens{X_{\mathrm{rot}}, \hat{X}_{\mathrm{rot}}},
\end{equation}
where \(\lambda_{\mathrm{rot}}\) is a hyperparameter to balance the two losses. We also design physical losses to constrain the physical aspects of the reconstructed motion. Similar to~\cite{humos,physdiff,tseng2023edge}, we include the floating loss \(L_{\mathrm{float}}\), the foot sliding loss \(L_{\mathrm{slide}}\), and the bone length loss \(L_{\mathrm{bone}}\) as geometric losses. \(L_{\mathrm{float}}\) minimizes the per-frame distance of the lowest joints above the ground from the ground plane. \(L_{\mathrm{slide}}\) minimizes the ground velocity of the foot joint if they are determined to be in ground contact using a distance threshold from the ground. \(L_{\mathrm{bone}}\) minimizes the deviation of bone lengths between the ground truth and the reconstructed motions. We write the final objective as
\begin{equation}
L_{\mathrm{vq}} = L_r + \lambda_f L_{\mathrm{float}} + \lambda_s L_{\mathrm{slide}} + \lambda_b L_{\mathrm{bone}},
\end{equation}
where \(\lambda_f\), \(\lambda_s\), and \(\lambda_b\) are the loss-balancing hyperparams.

\paragraph{Shape Data Augmentation.} To provide our SA-VAE with a wide diversity of subjects with different body shapes, beyond what is available in datasets, we adopt a data augmentation strategy. We replace \(q \%\) of the ground-truth shape parameters with synthetically generated ones, as mentioned in Section~\ref{sec:data}. Since there is no corresponding ground truth motion for the augmented shapes, we only optimize the indices with \(L_{\mathrm{slide}}\) and \(L_{\mathrm{bone}}\).

\subsection{Shape-Aware Motion Synthesis (\modelname)}
\label{sec:token_predict}

With the SA-VAE, we can tokenize a sequence of motion \(\bracks{x^N_1, x^N_2, \ldots,x^N_T}\) into a sequence of discrete indices \(\bracks{c_1, c_2, \ldots, c_\tau}\). Subsequently, our framework, \modelname, learns to map from language to shape-aware motion representations. 
We adopt a pretrained language model to predict the sequence of tokens.
We illustrate the overview of \modelname in Figure~\ref{fig:token}. To fine-tune the language model to understand the relationships between motion, shape, and text, we expand the original language model's vocabulary by adding \(k+2\) motion vocabulary items. Here, \(k\) is the number of codes from the quantizer codebook, and the additional two tokens are special tokens to signify the start and end of the motion token sequence. We also add one shape vocabulary token, \betatoken, to represent the shape. The language model takes text tokens as input and predicts tokens \(\braces{\text{\betatoken}, \hat{C}}\). We design the ground truth tokens to be \(\braces{\text{\betatoken}, C}\), \ie, we append the \betatoken~token with a sequence of motion tokens that align with the text input. We fine-tune the language model in an autoregressive manner.

Since shape parameters are fixed for a given body shape, unlike motion prediction, we cannot formulate shape parameter prediction as predicting a sequence of tokens. Representing shape would require an impractically large codebook, akin to using discrete numbers to fit a continuous space. To address this, previous works show that the last embedding in the language model can contain continuous information ~\cite{lisa,chatpose}. Instead of attempting to learn a discrete representation for shape, we extract the information through embedding. We denote the predicted output token as \(\hat{C_i}\), the corresponding embedding as \(M_i\), and the embedding corresponding to \betatoken~as \(M_\beta\). We obtain the shape parameter through a projector \(\mathrm{P}_{\theta_{e}}\), yielding \(\hat{\beta} = \mathrm{P}_{\theta_{e}}(M_\beta)\).

\paragraph{Objective.} 
To enable the language model to learn semantically consistent shape and motion vocabularies, we design the training data as mentioned in Section~\ref{sec:data}. The target outputs are the corresponding ground truth shape parameters and motion tokens. We train the model with the following objectives:
\begin{align}
L_{\mathrm{token}} &= \mathrm{CrossEntropy}\parens{C, \hat{C}} \\
L_{\mathrm{shape}} &= \lambda_{\beta} \modulus{\beta - \hat{\beta}},
\end{align}
where \(L_{\mathrm{token}}\) is the cross-entropy loss maximizing the log-likelihood of predicting the motion tokens, \(L_{\mathrm{shape}}\) is the \(L_1\) norm difference between the ground truth and the estimated shape parameters, and \(\lambda_{\beta}\) is the hyperparameter balancing the shape loss.

\section{Experiments}
\label{sec:exp}

In this section, we begin by discussing the experimental setup in Section~\ref{sec:exp_setup}. We then introduce the baselines used for comparison and present the results in Section~\ref{sec:exp_compare}, which includes quantitative comparisons, qualitative evaluations, and human perceptual assessments.

\subsection{Experimental Setup}
\label{sec:exp_setup}

\paragraph{Dataset.} 
We use the current largest text-to-motion dataset, HumanML3D~\cite{t2m}, which comprises 14,616 motion sequences and 44,970 motion descriptions. The motion sequences are sourced from 449 different subjects in AMASS~\cite{AMASS, AMASS_ACCAD, AMASS_HDM05, AMASS_TCDHands, AMASS_SFU, AMASS_BMLmovi, AMASS_CMU, AMASS_MoSh, AMASS_EyesJapanDataset, AMASS_BMLhandball, AMASS_PosePrior, AMASS_HumanEva, AMASS_DFaust, AMASS_TotalCapture} and HumanAct12~\cite{humanact12}. We follow the same data preprocessing procedures outlined for HumanML3D, but we omit normalizing the motions to a canonical model to get shape-aware motions, as detailed in Section~\ref{sec:data}.

\paragraph{Implementation Details.}
To train the SA-VAE, we configure the level \( \ell \) of the FSQ quantizer as \( \ell=[8,5,5,5] \), constructing a codebook that contains \( k=1000 \) indices with each index dimension set to 512. During training, we crop motion sequences to a length \( T = 64 \) and set the downsampled length \(\tau = 16\). We employ the AdamW optimizer with \( \beta_1=0.9 \) and \( \beta_2=0.99 \) (following standard AdamW notation --- no relation to the shape \(\beta\)), along with an exponential moving constant \( \lambda = 0.99 \). We train for 200K iterations at a learning rate of \( 2e^{-4} \), followed by 100K iterations at \( 1e^{-5} \). We set \(\lambda_f\), \(\lambda_s\), and \(\lambda_b\) to 10, and \(q\%\) for data augmentation to 10\%. With a batch size of 256, training takes approximately 12 hours on a single A100 GPU.

We use T5~\cite{t5} as our pretrained language model, which consists of 12 layers in both the transformer encoder and decoder. We employ AdamW optimizer with \( \beta_1=0.9 \) and \( \beta_2=0.99 \) at a learning rate of \( 8e^{-4} \). We set \(\lambda_{\beta}\) to 0.5 and train the model for 120K iterations on two distinct tasks: text-to-motion and motion-to-text. With a batch size of 64, this training phase requires one day on 8 A100 GPUs. Subsequently, we train the model solely on the text-to-motion task using the same batch size and learning rate for an additional 30K steps, which takes 10 hours on 8 A100 GPUs.

\begin{table}[t]
\centering
\caption{\textbf{Quantizer Reconstruction Comparison.} We report the reconstruction results, comparing our SA-VAE against baseline methods that utilize a VAE to quantize motion into discrete tokens. We assess the bone length difference, jitter score difference, and FID score relative to the shape-aware ground truth motions. Our VAE outperforms the baseline across all three metrics, particularly in reducing the bone length error by nearly half. These results demonstrate our model's effectiveness in aligning with the physical form of different body shapes. $\colorbox[RGB]{255,179,179}{\phantom{a}}$ and $\colorbox[RGB]{255,217,179}{\phantom{a}}$ highlight the \colorbox[RGB]{255,179,179}{best} and \colorbox[RGB]{255,217,179}{second-best} results.}
\resizebox{.8\columnwidth}{!}{
\begin{tabular}{lccc}
\toprule
\thead{Methods}  & \thead{FID$~\downarrow$} & \thead{Bone Length\\ Diff. (mm)$~\downarrow$} & \thead{Jitter\\ Diff. ($m/s^2$) $~\downarrow$} \\
\midrule
TM2T   ~\cite{tm2t}  & $0.528$ & $ 131.06$ & $ 79.87$ \\
MotionGPT ~\cite{motiongpt} & $0.173 $ & $ 94.27 $ & $ 40.44$ \\
T2M-GPT ~\cite{t2mgpt}  & \cellcolor{tabsecond}$0.151 $ & $\cellcolor{tabsecond} 83.42 $ & $ \cellcolor{tabsecond}34.75 $ \\
SA-VAE (Ours) &             $\cellcolor{tabfirst}{0.125} $ & $ \cellcolor{tabfirst}{45.88} $ & $ \cellcolor{tabfirst}{31.49}$ \\
\bottomrule
\end{tabular}
}
\label{tab:vae}
\end{table}


\begin{table}[t]
\centering
\caption{\textbf{Ablation Study.} \textit{sc} stands for shape-conditioning. $\colorbox[RGB]{255,179,179}{\phantom{a}}$, $\colorbox[RGB]{255,217,179}{\phantom{a}}$ and $\colorbox[RGB]{255,255,179}{\phantom{a}}$ highlight the \colorbox[RGB]{255,179,179}{best}, \colorbox[RGB]{255,217,179}{second-best} and \colorbox[RGB]{255,255,179}{third-best} results.}
\vspace{-3mm}
\resizebox{\columnwidth}{!}{
\begin{tabular}{lcccc}
\toprule
\thead{Methods}  & \thead{FID$~\downarrow$} & \thead{Bone Length\\ Diff. (mm)$~\downarrow$} & \thead{Float ($cm$) $~\downarrow$} & \thead{Skate ($\%$) $~\downarrow$} \\
\midrule
No sc                                   & 0.148 & 99.18  & 0.575 &  \cellcolor{tabthird}6.76  \\
sc                                      & \cellcolor{tabfirst}0.105 & 66.41   & 0.567 & \cellcolor{tabsecond}6.60 \\
sc$+L_{bone}$                           & \cellcolor{tabsecond}0.107 & \cellcolor{tabfirst}45.11   & \cellcolor{tabthird}0.480 & 7.07 \\
sc$+L_{bone}+L_{float}$                 & 0.137 & \cellcolor{tabthird}45.97   & \cellcolor{tabfirst}0.255 & 7.90 \\
sc$+L_{bone}+L_{float}+L_{skate}$ (Full) & \cellcolor{tabthird}0.125 & \cellcolor{tabsecond}45.88  & \cellcolor{tabsecond}0.266 & \cellcolor{tabfirst}6.14 \\
\bottomrule
\end{tabular}
}
\label{tab:vae_ablation}
\end{table}

\begin{table}[t]
\centering
\caption{\textbf{Attributes Prediction.} We present the differences $(cm)$ across six attributes between our beta predictions and the ground truth, focusing solely on our method as no comparable works predict beta concurrently. Our model demonstrates a robust ability to predict the correct beta values, with discrepancies from the ground truth around one $cm$. \textit{C.} stands for circumference.}
\resizebox{\columnwidth}{!}{
\begin{tabular}{lcL{0.3cm}lcL{0.3cm}lc}
\toprule
Attributes & Difference && Attributes & Difference && Attributes & Difference \\
\cmidrule{1-2}\cmidrule{4-5}\cmidrule{7-8}
Height & $0.5767$ && Arm Length & $0.0433$ && Leg Length & $0.0505$ \\
Chest C. & $0.6921$ && Waist C. & $1.0558$ && Hip C. & $0.6044$ \\
\bottomrule
\end{tabular}
}
\label{tab:attribute}
\end{table}

\paragraph{Evaluation Metrics.}
To evaluate the physical plausibility of motions, we adopt the physics-based metrics of~\cite{humos, physdiff}. The \textit{Penetrate} metric measures ground penetration by computing the distance between the ground and the lowest body joints below the ground. \textit{Float} measures the distance between the ground and the lowest body joints above the ground. \textit{Skate} measures foot skating by computing the percentage of frames where the foot joints in contact with the ground have a velocity over a threshold. We also propose a new metric, \textit{Bone Length Variances}, which measures the variance of arm and leg lengths across the entire sequence to evaluate the stability of the generated motion. Following HUMOS~\cite{humos}, we exclude motions involving climbing and compute the \textit{Penetrate}, \textit{Float}, and \textit{Skate} on the remaining dataset. This selection ensures the accuracy and relevance of the metrics for motions primarily characterized by ground-based dynamics.

We follow the evaluation protocol proposed by Guo \etal~\cite{t2m} to evaluate text-motion alignment and motion quality. \textit{Fr\'echet Inception Distance (FID)} evaluates the distance of feature distributions between the generated and real motions. \textit{R-Precision} evaluates the accuracy of matching between texts and motions using the top 1/2/3 retrieval accuracy. \textit{Multimodal Distance (MM-Dist)} measures the feature distance between motions and texts. \textit{Diversity} calculates the variance through features extracted from the motions. We compute all results with a 95\% confidence interval obtained from 20 repeated runs.

\begin{figure*}[t]
\centering
\mpage{0.19}{\small{Ground Truth}}\hfill
\mpage{0.19}{\small{Ours}}\hfill
\mpage{0.19}{\small{T2M-GPT~\cite{t2mgpt}}}\hfill
\mpage{0.19}{\small{MotionGPT~\cite{motiongpt}}}\hfill
\mpage{0.19}{\small{MotionDiffuse~\cite{motiondiffuse}}}\\
\mpage{1.0}{\includegraphics[width=\linewidth, trim=0 0 0 0, clip]{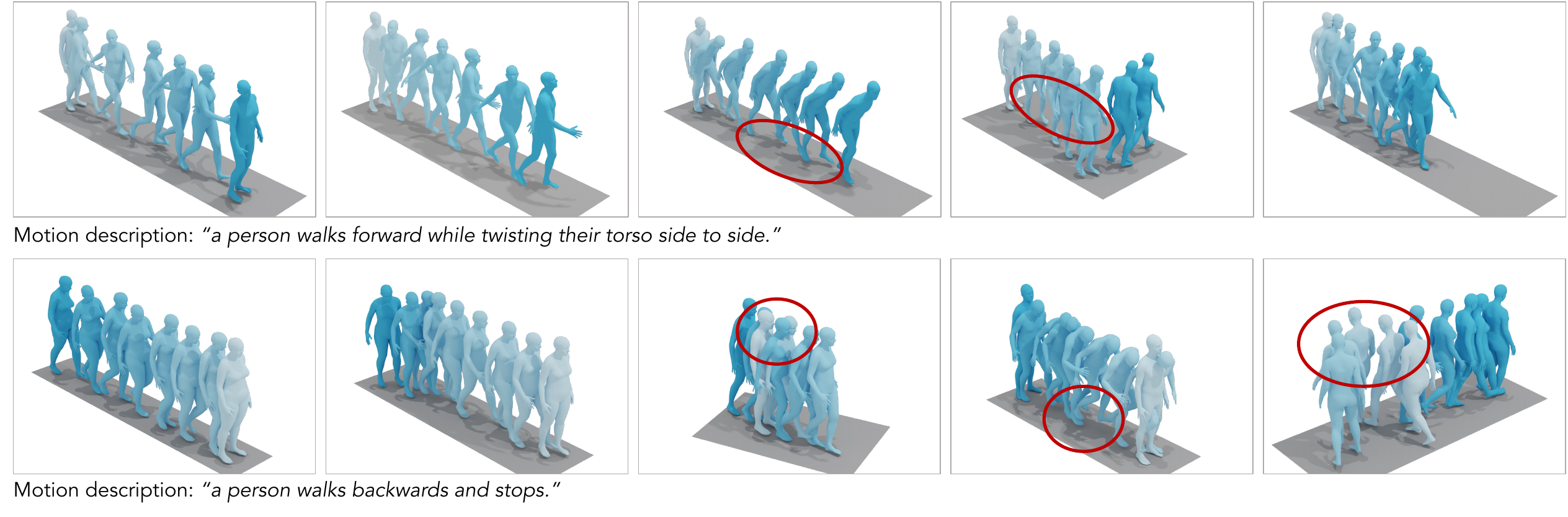}}\\
\caption{\textbf{Qualitative Comparisons.} We compare our method with three baseline methods, T2M-GPT~\cite{t2mgpt}, MotionGPT~\cite{motiongpt}, and MotionDiffuse~\cite{motiondiffuse}, illustrating two samples from the HumanML3D test set. The motions are colored from light to dark blue to represent progression over time. We highlight issues such as incorrect foot motion and other inaccuracies that do not align with expected motion patterns. 
Our method not only generates motions that align with the textual descriptions, but also accurately follows the body attributes and physical dynamics of the ground truth. Additional visual results and detailed comparisons are available in the project website.
}
\label{fig:qualitative}
\end{figure*}


\begin{figure*}[t]
\centering
\mpage{0.325}{\includegraphics[width=\linewidth, trim=0 1.2cm 0 0, clip]{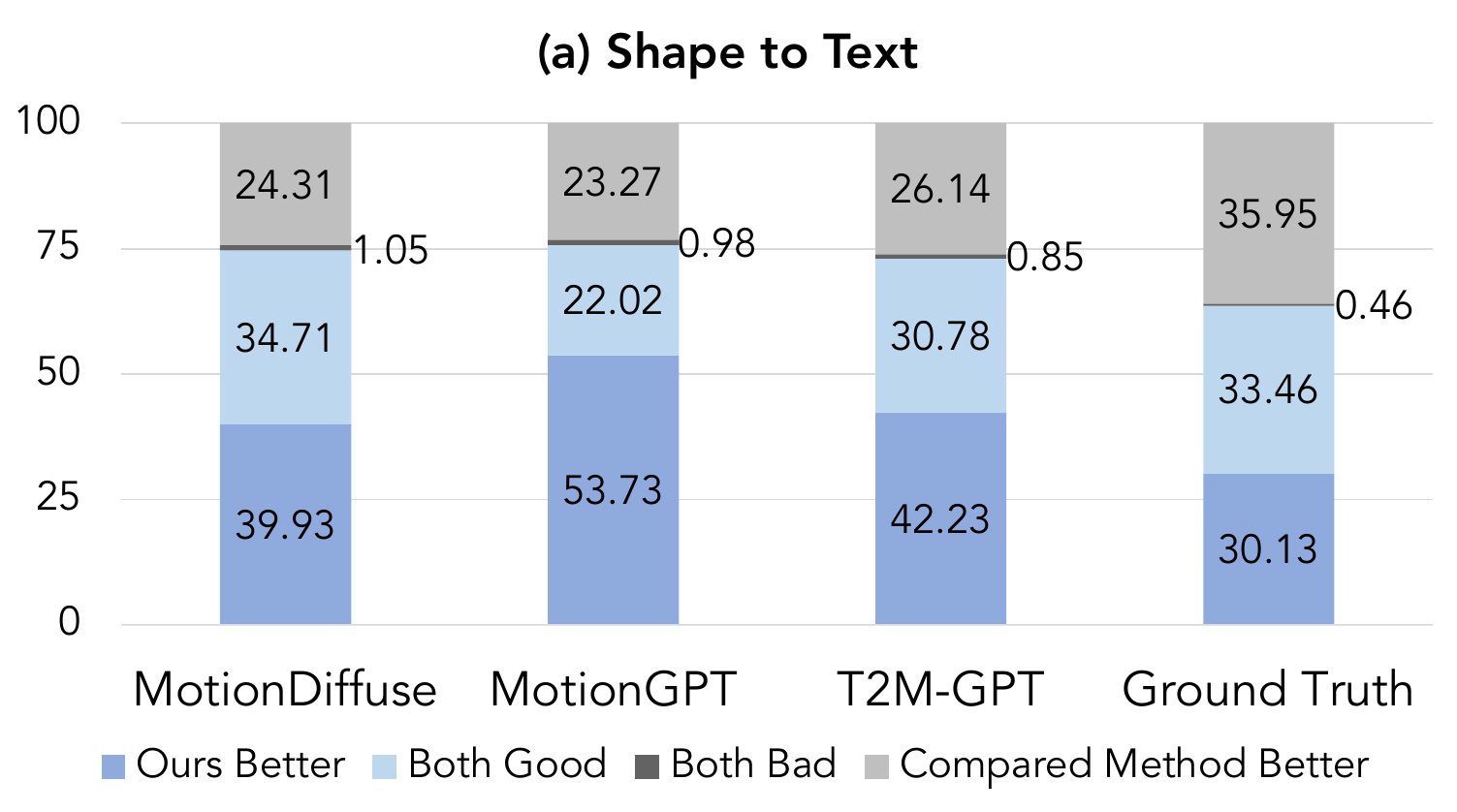}}\hfill
\mpage{0.325}{\includegraphics[width=\linewidth, trim=0 1.2cm 0 0, clip]{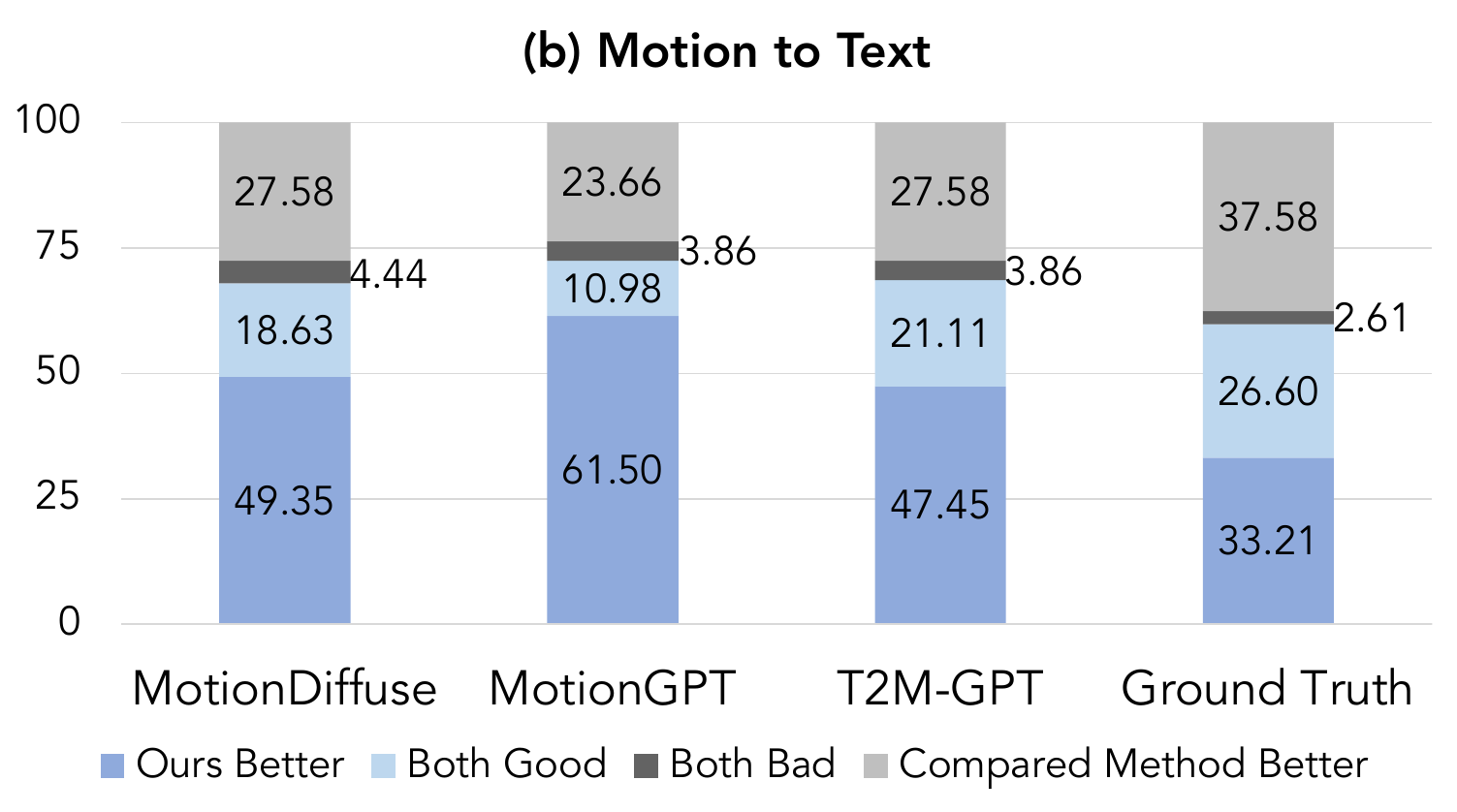}}\hfill
\mpage{0.325}{\includegraphics[width=\linewidth, trim=0 1.2cm 0 0, clip]{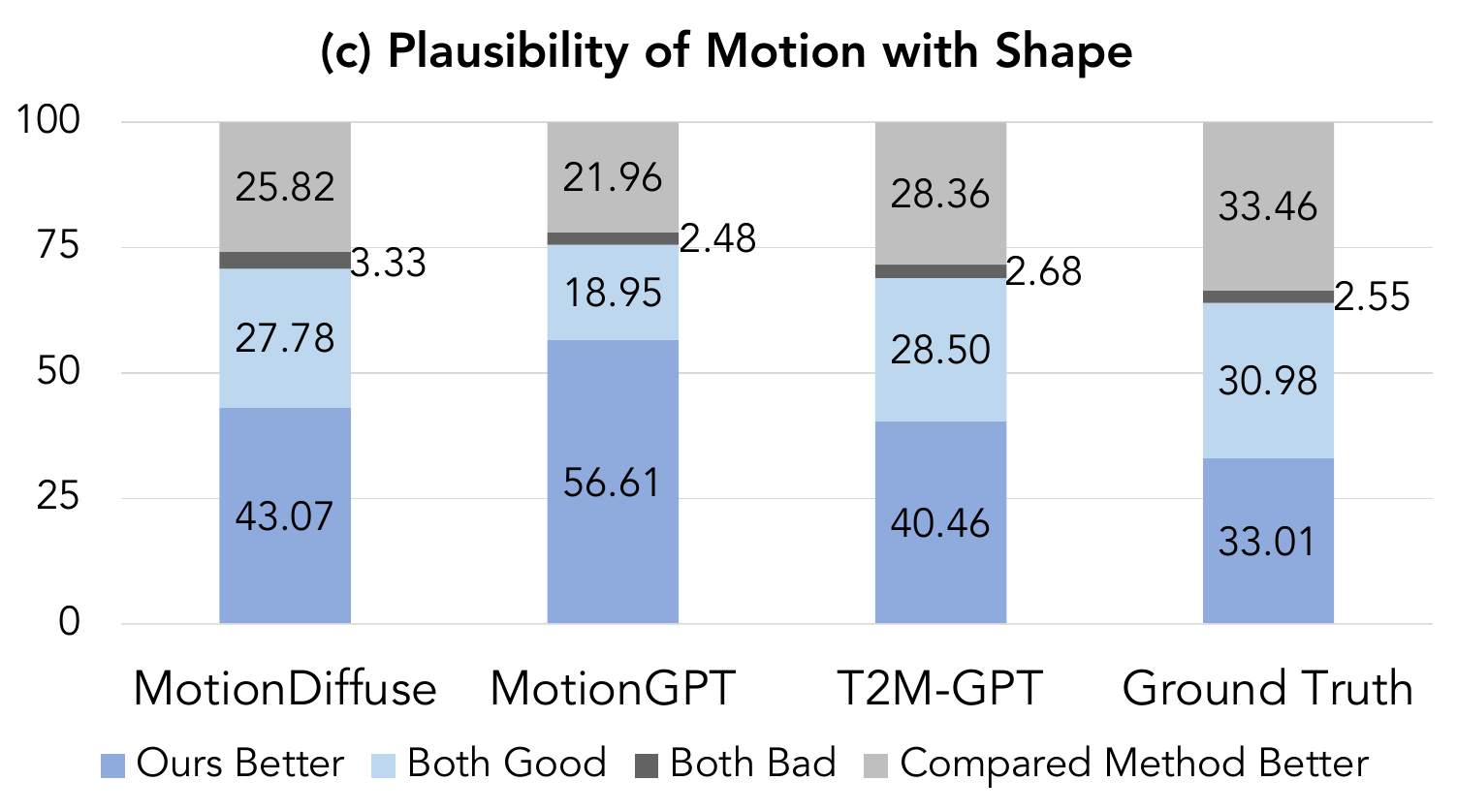}}\\

\mpage{1.0}{\includegraphics[width=.5\linewidth, trim=0 0cm 0cm 13cm, clip]{images/userstudy/plau_motion.pdf}}\\

\caption{\textbf{Perceptual Evaluation.} We show the distributions of aggregate responses from annotators on their preferences for samples generated by our method and baseline methods, including MotionDiffuse~\cite{motiondiffuse}, MotionGPT~\cite{motiongpt}, and T2M-GPT~\cite{t2mgpt}, as well as the corresponding ground truth samples. We assess the distributions on three metrics: \textit{(a) Shape to Text}, how well the body shape matches the text input; \textit{(b) Motion to Text}, how well the motion matches the text input; and \textit{(c) Plausibility of Motion with Shape}, how realistic the motions appear for the corresponding body shapes. Across all three metrics, we observe that our method is preferred nearly as much as the ground truth and is favored by approximately 12\% to 38\% over the baselines.
}
\label{fig:userstudy}
\end{figure*}

\subsection{Comparison with Baselines}
\label{sec:exp_compare}

We compare our results with the existing state-of-the-art approaches, including T2M \cite{t2m}, TM2T \cite{tm2t}, MLD \cite{mld}, MotionDiffuse \cite{motiondiffuse}, MDM \cite{mdm}, T2M-GPT \cite{t2mgpt}, and MotionGPT \cite{motiongpt}. We train these baseline methods using our training data, including shape and motion descriptions, and task the models to predict shape-aware motions. We note that T2M \cite{t2m} and TM2T \cite{tm2t} employ a specific vocabulary for text encoder training and thus cannot incorporate shape descriptions as input. Therefore, we only input motion descriptions for these models and report the results as a reference. In the following subsections, we present quantitative, qualitative, and perceptual evaluation results, demonstrating that our model achieves state-of-the-art performance across these metrics.

\paragraph{Quantitative Results.}
\label{sec:exp_quan}

We present the physical plausibility comparison in Table~\ref{tab:physic_text2motion}. Our method outperforms baseline models that share the same settings, achieving the best performance in the Penetrate, Skating, and Bone Length metrics and ranking second best in the Float metric. This demonstrates our method's effectiveness in learning physical rules. 
The metrics might not fully reflect the true realism of the generated motion for all motion scenarios. Hence, we also conduct a perception evaluation with human participants for a comprehensive assessment.

We also report results on text-motion alignment metrics in Table~\ref{tab:physic_text2motion}, where our model outperforms the same set of baselines across all metrics. Our model also achieves the best FID score over baselines, even outperforming baselines that require ground truth motion length during generation (which has limited practicality in real-world applications).

We report the quantizer performance in Table~\ref{tab:vae}. TM2T~\cite{tm2t}, T2M-GPT~\cite{t2mgpt}, and MotionGPT~\cite{motiongpt}, all quantize motions into discrete tokens. We present the performance of our SA-VAE in comparison with these models. From Table~\ref{tab:vae}, we focus on the reconstruction quality of the model, emphasizing three metrics: the FID score to assess the motion quality, the bone length difference to measure the skeletal discrepancies between the reconstructed and ground truth motions, and the jitter difference to evaluate the disparity in the mean acceleration between reconstructed and ground truth motions. Our model demonstrates superior performance on all three metrics, achieving the lowest FID score and approximately half the bone length difference, which underscores the effectiveness of our SA-VAE in reconstructing shape-aware motions.

As we are the first, to the best of our knowledge, to predict shape parameters concurrently, we exclusively report the differences in body attributes. Table~\ref{tab:attribute} presents these results, showing that our model can predict shape parameters differing by around 1 \(cm\) from the ground truth.

We also conduct an ablation study to evaluate the impact of the various components used in our model, including \textit{shape-conditioning}, bone length loss \(L_{\mathrm{bone}}\), float loss \(L_{\mathrm{float}}\), and foot skate loss \(L_{\mathrm{skate}}\). We assess these variants on FID, bone length difference, floating, and foot skate ratio. Table~\ref{tab:vae_ablation} shows how each component improves performance across the metrics.
We select our final model configurations as the one achieving the best overall results.

\paragraph{Qualitative Results.}
\label{sec:exp_qual}
Figure~\ref{fig:qualitative} illustrates a qualitative comparison on two samples from the HumanML3D test set. Our method can generate realistic motions that are well-aligned with both shape and motion descriptions. In contrast, body shapes in other methods are obtained indirectly via fitting the SMPL model to the generated joints. The red circles highlight unrealistic motions, such as floating feet, misaligned postures, incomplete actions, self-intersections, and deviating from the prompt. More visual results are available on the website.

\paragraph{Perceptual Evaluation.}
\label{sec:exp_user}
We conducted a human perceptual evaluation with pairwise comparisons between motion samples generated by our method and other baselines, as well as the corresponding ground truth samples.
We selected the three best baselines from our quantitative evaluations, which also follow the same generation paradigm as ours, for comparison: MotionDiffuse~\cite{motiondiffuse}, MotionGPT~\cite{motiongpt}, and T2M-GPT~\cite{t2mgpt}. For each pairwise comparison, we show our sample and the other sample (one of the baselines or ground truth), for the same text description, in a random order to the participants. We ask them to rank the two samples as one of \textit{sample 1 better}, \textit{sample 2 better}, \textit{both similarly good} or \textit{both similarly bad} on three metrics: shape-to-text accuracy, motion-to-text accuracy, and plausibility of motion with shape.
We randomly selected 34 text descriptions from the HumanML3D test set to generate the sample sets for the perceptual study, and recruited participants through Amazon Mechanical Turk. After filtering out inadmissible study responses (\eg, responses that were too quick and self-inconsistent), we report our perceptual evaluation results with 45 respondents in Figure~\ref{fig:userstudy}. We note that our method was consistently preferred over the baselines and even matched the preference rate of the ground truth.

\section{Limitations}
\label{sec:limitations}


One limitation of our approach is the constrained nature of the shape descriptions used for motion generation. The preprocessing step requires descriptions to adhere to a specific template, which may restrict the model's ability to handle various natural language inputs and diverse body shapes. This limitation could be mitigated by incorporating a more diverse dataset with a broader range of descriptive styles. Additionally, employing an advanced language model to preprocess and standardize descriptions to fit the required template can enhance the model's versatility and real-world applicability.

\section{Conclusions}
\label{sec:conc}

We have presented a novel framework for shape-aware motion synthesis from textual descriptions, with the ability to generate both shape parameters and shape-aware motions from input text descriptions. 
Our method effectively captures the diverse correlations between human body shapes and motion dynamics, as evidenced by comprehensive evaluations that include quantitative results, qualitative comparisons, and perceptual studies with humans.

{
    \small
    \bibliographystyle{ieeenat_fullname}
    \bibliography{main}
}


\end{document}